# Lensless-camera based machine learning for image classification

GANGHUN KIM,[1] STEFAN KAPETANOVIC,[1] RACHAEL PALMER,[2] RAJESH MENON,[*]

[1]Department of Electrical and Computer Engineering, University of Utah, Salt Lake City, UT 84112
[2]Department of BioEngineering, University of Utah, Salt Lake City, UT 84112
*Corresponding author: rmenon@eng.utah.edu

**Machine learning (ML) has been widely applied to image classification. Here, we extend this application to data generated by a camera comprised of only a standard CMOS image sensor with no lens. We first created a database of lensless images of handwritten digits. Then, we trained a ML algorithm on this dataset. Finally, we demonstrated that the trained ML algorithm is able to classify the digits with accuracy as high as 99% for 2 digits. Our approach clearly demonstrates the potential for non-human cameras in machine-based decision-making scenarios.**

*OCIS codes: (150.0150) Machine Vision; (100.3190) Inverse Problems; (110.1758) Computational Imaging.*

Recently, it has become common to train machine-learning (ML) algorithms to recognize objects in images by exposing them to vast databases of labeled images [1-4]. Spectacular gains in classification accuracy have been obtained and ML algorithms are now able to make complex decisions based upon object recognition in image and video data. These algorithms are typically educated on conventional (what we refer to as human-centric) images. Recently, there have also been significant advances in lensless imaging, where a sensor that does not have a lens captures information from a scene or object [5-7]. Such lensless cameras offer advantages of simplicity, low cost, reduced weight and small form factors. Previously, we demonstrated that frames captured by a bare CMOS sensor can be used to reconstruct images that are recognizable by humans. Here, we show that such human-centric reconstruction is actually not necessary for machine-based image recognition and classification. Specifically, we train a ML algorithm on a database of frames created by a lensless camera and demonstrate that the algorithm is capable of image classification using the data directly from the lensless camera.

Our lensless camera is simply a conventional bare CMOS sensor (The Imaging Source, DFM 22BUC03-ML). We use a liquid-crystal display (LCD) to show various images of handwritten digits from 0 to 9 [8]. Sample images of handwritten digits were obtained from the MNIST database. MNIST database has been used for image classification test since the 1990's, and is widely used for assessing image classification accuracy. The LCD is placed about 250mm away from and facing the CMOS sensor as shown in Fig. 1(a). The exposure time is 150ms and we average 100 frames to reduce noise. Using this procedure, we created a database of 70,000 images with appropriate labels. Examples of the original MNIST images and their corresponding lensless images are shown in Fig. 1(b).

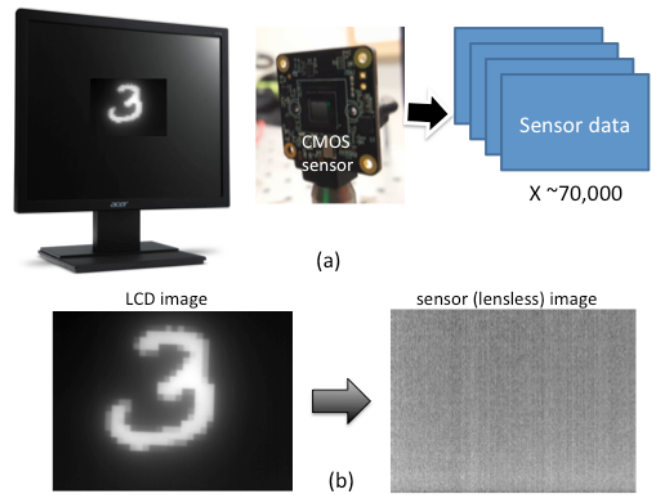

Figure 1: (a) Schematic of our experiment. Handwritten digits are displayed on an LCD, and lensless images are captured by a bare CMOS sensor. Approximately 70,000 training images were created. (b) Example of one handwritten digit ("3") and its corresponding lensless image.

As raw images are fairly large (307,200 pixels in each image), it is not suitable for large scale training in its raw form. Hence, SURF (speeded up robust features) extraction [9] and subsequent K-mean clustering [10] are performed to reduce data dimension. Note that we used 2000 to 3000 images to perform feature extraction as they provide sufficient information to approximate image features that are present throughout the whole set. Extracted features are then piped to the Matlab machine learning toolbox (also called classification learner) [11], after data is properly formatted for the tool, to perform the training. Subsequently, we used a variety of learning methods to train the model. Algorithms used for the training include support vector machine (SVM) [1], decision tree [12], k-nearest neighbors (KNN) [3], and their variants. Once trained, each trained model is measured for accuracy using set number of test images as described below. We then picked the model with the highest prediction accuracy, which we report in Fig. 2. In most cases, the SVM algorithm achieved the highest accuracy.

Next, we also generated lensless images in the same manner for testing the trained algorithm. We ensured that the images used for testing were completely different than the ones used during feature extraction and ML training. The training and testing procedures were separately conducted for databases comprising of 2 digits ("0" and "1"), 5 digits ("0" to "") and 10 digits ("0" to "9"). The accuracy with which the ML algorithm classifies the lensless frames into the appropriate digit class was verified against the known image labels, allowing us to measure the classification accuracy. The classification accuracy is plotted for the 3 cases (2, 5 and 10 digits) as a function of the number of training images in Fig. 2. As the number of training images increases, the classification accuracy increases and seems to saturate for all cases. The classification accuracy decreases as the number of digits increases. The best classification accuracy is about 99% for 2 digits with 7,800 training images. Although the classification accuracy is lower for more number of digits, this can be improved by using deep convolutional networks or related algorithms.

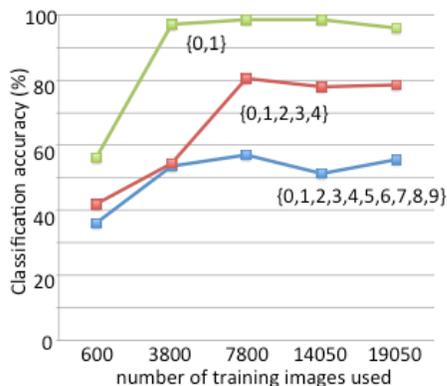

Figure 2: Best classification accuracy as a function of number of training images used.

In conclusion, we demonstrate that machine learning can be effectively applied to classify lensless images. Such non-human image-based decision-making could lead to significant improvements in the ability of autonomous agents to navigate and make sense of the external world.

**Funding.** National Science Foundation (Grant # 10037833). Utah Science Technology and Research Initiative. University of Utah's Undergraduate Research Opportunities Program (UROP).